\documentclass[11pt]{article}

\usepackage[preprint]{acl}

\usepackage{times}
\usepackage{latexsym}

\usepackage[T1]{fontenc}

\usepackage[utf8]{inputenc}

\usepackage{microtype}

\usepackage{inconsolata}

\usepackage{graphicx}

\usepackage{algorithm}
\usepackage{algorithmic}
\usepackage{float}
\usepackage{booktabs} 
\usepackage{makecell}  
\usepackage[table]{xcolor}  
\usepackage{tabularx}
\usepackage{multirow}
\usepackage{enumitem} 
\usepackage{amsmath}
\usepackage{amssymb}
\usepackage[dvipsnames]{xcolor}
\usepackage[most]{tcolorbox}
\newtcolorbox{promptbox}[1][]{
    colback=gray!10,      
    colframe=Blue!20!black,    
    boxrule=0.5mm,        
    arc=1mm,              
    boxsep=0mm,
    fontupper=\ttfamily\small, 
    width=0.5\textwidth,     
    title=#1,
    fonttitle=\ttfamily\footnotesize\centering,
}

\setlist{%
  topsep=4pt,    
  partopsep=0pt, 
  itemsep=1pt,   
  parsep=0pt     
}

\title{Chart-RL: Generalized Chart Comprehension via Reinforcement Learning with Verifiable Rewards}

\author{
 \textbf{Xin Zhang},
 \textbf{Xingyu Li},
 \textbf{Rongguang Wang},
 \textbf{Ruizhong Miao},
\\
 \textbf{Zheng Wang},
 \textbf{Dan Roth},
 \textbf{Chenyang Li}
\\
\\
 Oracle AI
\\
\\
 \small{
   \{xin.j.zhang, xingyu.li, rongguang.wang, ruizhong.miao, z.zheng.wang, dan.roth, chenyang.li\}@oracle.com
 }
}

\begin{document}
\maketitle

\begin{abstract}
Accurate chart comprehension represents a critical challenge in advancing multimodal learning systems, as extensive information is compressed into structured visual representations. However, existing vision-language models (VLMs) frequently struggle to generalize on unseen charts because it requires abstract, symbolic, and quantitative reasoning over structured visual representations. In this work, we introduce Chart-RL, an effective reinforcement learning (RL) method that employs mathematically verifiable rewards to enhance chart question answering in VLMs. Our experiments demonstrate that Chart-RL consistently outperforms supervised fine-tuning (SFT) across different chart understanding benchmarks, achieving relative improvements of 16.7\% on MutlChartQA, and 11.5\% on ChartInsights. We conduct robustness analysis, where Chart-RL achieves enhanced performance in 18 of 25 perturbed chart categories, demonstrating strong consistency and reasoning capability across visual variations. Furthermore, we demonstrate that task difficulty and inherent complexity are more critical than data quantity in RL training. For instance, Chart-RL trained on merely 10 complex chart-query examples significantly outperforms models trained on over 6,000 simple examples. Additionally, training on challenging reasoning tasks not only improves in-domain generalization relative to simpler tasks, but also facilitate strong transfer to out-of-domain visual mathematical problems. 
\end{abstract}
\section{Introduction}
Reinforcement learning with verifiable rewards (RLVR) has demonstrated remarkable success in enhancing the reasoning capabilities of LLMs~\citep{guo2025deepseek, jaech2024openai, team2023gemini}. However, one critical area that remains relatively underexplored is chart understanding in VLMs, whose challenges such as symbolic reasoning and precise data extraction distinguish it fundamentally from traditional vision-language tasks. Unlike natural images, charts exhibit extreme diversity in types (bar charts, pie charts, scatter plots, etc.) and structural organizations, with performance heavily dependent on the training data distribution \citep{huang2024pixels}. Current VLMs often struggle to bridge the gap between extracting descriptive information and deriving correct answers that require multi-step reasoning \citep{kamath2023s}. 

\begin{figure}[!ht]
    \centering
    \includegraphics[width=0.9\linewidth]{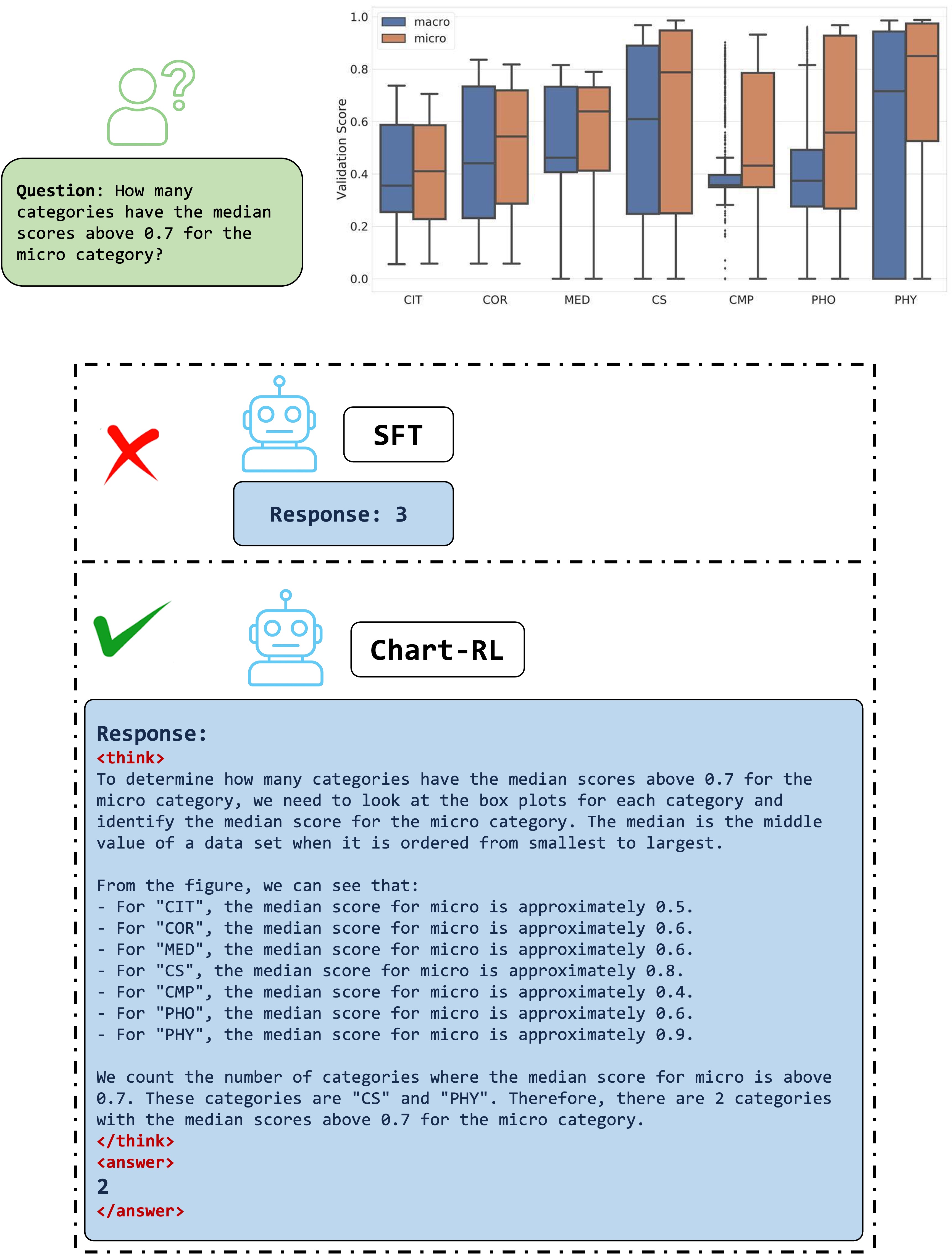}
    \caption{Comparison of Chart-RL vs SFT to reason on complex chart question answering.}
    \label{fig:sft-vs-Chart-RL}
\end{figure}

Recent studies to improve chart question answering have explored various strategies, such as converting charts into tabular formats suitable for prompting LLMs~\citep{liu2022deplot}, instruction tuning on multiple chart-related objectives ~\citep{schleid2025visual}, and generating synthetic chart data to augment VLM's capability in real-world chart comprehension ~\citep{huang2025evochart}. Nonetheless, the majority of chart comprehension methods rely on either larger curated datasets that cover broader chart and question types, or SFT using domain-specific datasets. While large-scale datasets such as ChartQA ~\citep{masry2022chartqa}, PlotQA ~\citep{methani2020plotqa}, and FigureQA ~\citep{kahou2017figureqa} have significantly expanded coverage of chart types and reasoning scenarios, they often focus on specific chart formats with template-based questions that may not fully capture real-world complexity. Although domain-specific SFT can enhance model performance on targeted task categories, it frequently leads to catastrophic forgetting and performance degradation on untrained tasks due to data distribution shifts ~\citep{luo2023empirical, kalajdzievski2024scaling}. This phenomenon fundamentally limits the capability of current approaches to generalize across diverse chart types and reasoning scenarios in real-world applications ~\citep{schulz2025testing, huang2023lvlms}.

To address these limitations, we propose \textbf{Chart-RL}, an efficient reinforcement fine-tuning framework that employs R1-style reinforcement learning to enhance chart understanding capabilities in VLMs. A key insight motivating our approach is that chart-based questions frequently correspond to mathematically verifiable ground truths, making them well-suited for rule-based reward functions in RL training. More importantly, many charts encapsulate sufficiently complex visual and semantic information that necessitate iterative reasoning processes. We find that such reasoning complexity is of central importance because it enables models to develop advanced reasoning capabilities that can't be acquired through simple chart interpretation tasks ~\citep{parashar2025curriculum}. Leveraging group relative policy optimization (GRPO) with chart-aligned accuracy and format rewards \citep{shao2024deepseekmath, shen2025vlm}, we train VLMs to develop robust reasoning capabilities that generalize across different chart types and complexity levels. Our approach circumvents the limitation of conventional SFT methods, where task-specific adaptation results in narrow specialization rather than transferable reasoning skills \citep{luo2023empirical}.

As illustrated in Figure ~\ref{fig:sft-vs-Chart-RL}, we demonstrate that Chart-RL improves performance across different chart understanding tasks, ranging from simple data extraction to complex multi-step reasoning, while achieving remarkable data efficiency. By training on complex chart problems, Chart-RL develops proficiency in low-level component tasks without requiring explicit instruction tuning. 

The key contributions of this paper include:

\begin{itemize}
    \item \textbf{First Adoption of RLVR for Chart Comprehension in VLMs:} We introduce Chart-RL, a reinforcement fine-tuning strategy that uses reinforcement learning with verifiable rewards to enhance chart comprehension capabilities in VLMs, achieving significant improvements over SFT approaches.
    
    \item \textbf{Efficient Generalization with Superior Training Data Efficiency:} We demonstrate, for the first time, that RL training on complex chart reasoning tasks enables robust generalization, achieving consistent improvements across diverse benchmarks without task-specific optimization.
    
    \item \textbf{Comprehensive Experimental Validation of Robustness and Generalization:} Our extensive empirical evaluation reveals a critical insight: \textbf{task complexity}, rather than the quantity of training data, is pivotal in developing generalizable chart understanding capabilities.
\end{itemize}

\begin{figure*}[htbp]
    \centering
    \includegraphics[width=1.0\textwidth]{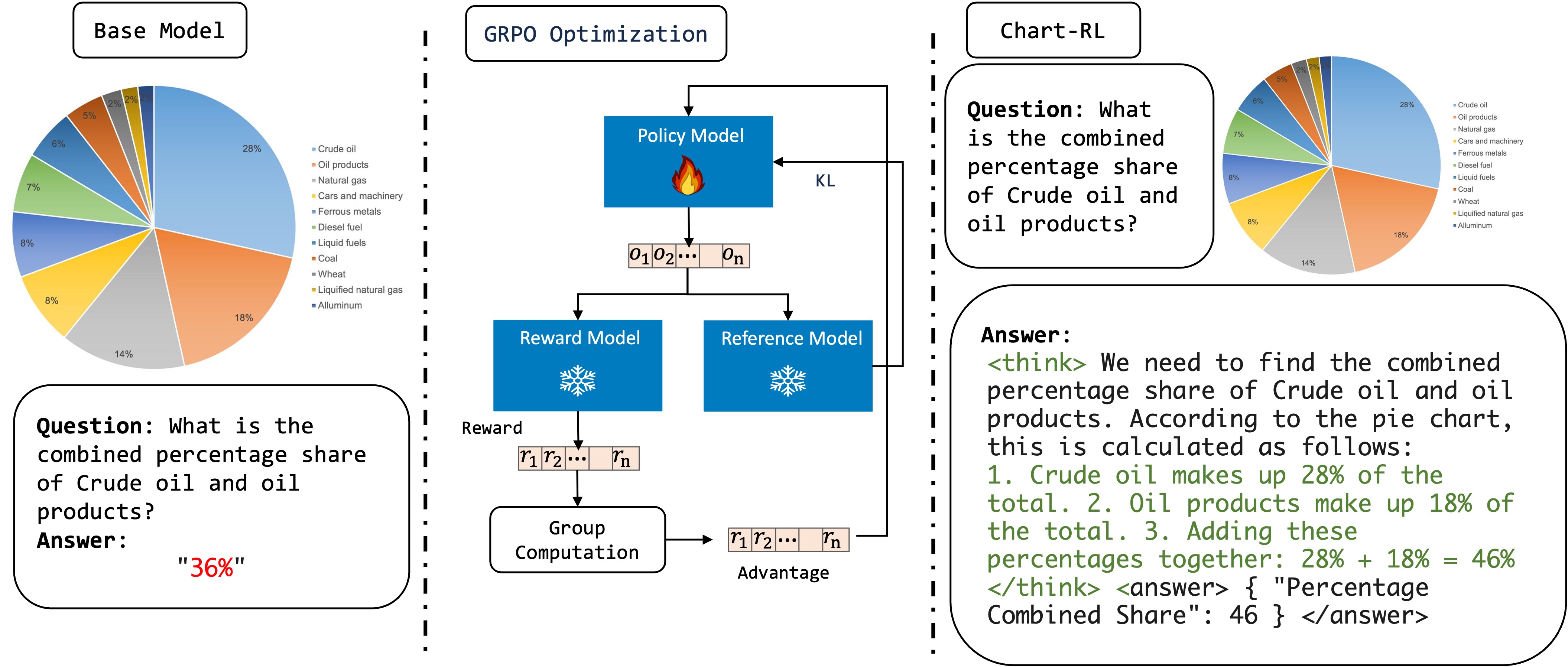}
    \caption{Schematic diagram of Chart-RL framework with GRPO optimization for chart comprehension.}
    \label{fig:Chart-RL-framework}
\end{figure*}
\section{Related Works}
\subsection{Chart-Based Question Answering}
Chart understanding has evolved from multi-stage pipelines \citep{masry2022chartqa} to end-to-end models. This transformation was pioneered by OCR-free architectures like Donut \citep{kim2022ocr-free}, which paved the way for powerful VLMs. Recent progress has been predominantly achieved through SFT approaches, with models such as Pix2Struct \cite{lee2023pix2struct} and UniChart \citep{masry2023unichart} being adapted for direct chart-to-answer generation. Concurrently, advanced models including LLaVA \cite{liu2023visual}, PaliGemma \citep{beyer2024paligemma}, and ChartLlama \citep{han2023chartllama} have been fine-tuned to address more sophisticated and diverse reasoning challenges. 

Despite its success in task-specific adaptation, SFT faces critical limitations in the domain of chart understanding. The need for large, meticulously annotated datasets makes the process costly and prone to overfitting, where models learn to exploit data artifacts rather than acquiring robust reasoning principles. Furthermore, SFT-based models often exhibit poor generalization ability: performance gains on one type of task do not reliably transfer to related tasks or different data distributions \cite{schulz2025testing}. 

\subsection{Reinforcement Fine-Tuning in VLMs}
The success of R1-style reinforcement learning in enhancing LLM reasoning capabilities \citep{jaech2024openai, guo2025deepseek} has inspired extensive research into its application toVLMs. For example, R1-OneVision \citep{yang2025r1} and Vision-R1 \citep{huang2025vision} construct multimodal reasoning datasets by converting visual information into textual format before RL training, while other studies focus on specific applications such as object counting \citep{chen2025r1} or employ staged training strategies with progressive difficulty rewards \citep{peng2025lmm, deng2025boosting}. Notably, several studies have observed the ``visual aha moment'' in VLMs \citep{zhou2025r1, meng2025mm}, demonstrating emergent reasoning capabilities. The VLM-R1 framework \citep{shen2025vlm} provides a comprehensive investigation into applying RL to general visual understanding tasks, including Referring Expression Comprehension (REC) and Open-Vocabulary Object Detection (OVD).

Despite these advances, the reasoning capabilities cultivated by existing R1-style VLMs are primarily tailored for understanding of natural images. Since charts present highly condensed information in diverse structural formats, the unique challenges of chart understanding have yet to be systematically addressed. These tasks require interpreting relationships, performing calculations, and synthesizing information from graphical elements like bars, lines, and legends. Building on the principles established in the VLM-R1 \citep{shen2025vlm}, we propose an R1-type reinforcement fine-tuning approach specifically for chart understanding, aiming to instill the robust generalization and reasoning capabilities.

\section{Methods}
\subsection{Chart-RL Framework}
We present Chart-RL, an extension of the open-source VLM-R1 platform \citep{shen2025vlm} that adapts R1-style reinforcement fine-tuning for chart understanding tasks. As shown in Figure ~\ref{fig:Chart-RL-framework}, we implement a verifiable reward function that produces a score based on alignment between model outputs and annotated ground truths. In addition, a binary format reward function is employed to enforce task-specific output structures. This modular reward design enables Chart-RL to adapt to different chart-related training tasks with mathematically verifiable annotations.  

\subsection{Accuracy and Format Rewards} 

In Chart-RL, we employ the GRPO algorithm, which evaluates a group of candidate responses to guide policy updates, eliminating the need for a separate critic \citep{shao2024deepseekmath}. The GRPO algorithm begins by sampling $N$ candidate responses $\{o_1, o_2, ..., o_N\}$ from the current policy $\pi_{\theta}$ for a given query $q$. Each response $o_i$ is scored using a reward function $R(q, o_i)$, and the resulting rewards $r_i$ are normalized to compute the advantage $A_i$ for each response as: 
\begin{equation}\label{eq:grpo_reward}
    A_{i} = \frac{r_{i} - \text{mean}\{r_{1}, r_{2}, ..., r_{N}\}}{\text{std}\{r_{1}, r_{2}, ..., r_{N}\}},
\end{equation}
where $A_i$ measures the relative advantage of response $i$ compared to other responses in the sampled group.
The policy $\pi_{\theta}$ is then updated to increase the likelihood of generating responses with higher advantages using the following objective:
\begin{equation} \label{eq:grpo_objective}
\begin{aligned}
    \mathcal{J}_{GRPO}(\theta) & = \mathbb{E}_{[\{o_{i}\} \sim \pi_{\theta_{old}}(q)]} \\
     \frac{1}{N}\sum_{i=1}^{N} & \left\{ \min[s_{1} \cdot A_{i}, s_{2} \cdot A_{i}] - \beta \mathbb{D}_{KL}[\pi_{\theta} || \pi_{\text{ref}}] \right\} \\
    s_{1} = & \frac{\pi_{\theta}(o_{i}|q)}{\pi_{\theta_{old}}(o_{i}|q)} \\
    s_{2} = & \text{clip}\left(\frac{\pi_{\theta}(o_{i}|q)}{\pi_{\theta_{old}}(o_i|q)}, 1-\epsilon, 1+\epsilon\right)
\end{aligned}
\end{equation}

\paragraph{Accuracy Reward.} To ensure verifiable training signals, we sample query-chart pairs that produce mathematically deterministic results for RL training. Our accuracy reward design is based on the relative error between the model's predicted result $v_{\text{p}}$ and the ground-truth solution $v_{\text{g}}$, evaluated against a predefined precision threshold $\tau$. The accuracy reward score $S(v_{\text{p}}, v_{\text{g}})$ is computed as:

\begin{equation} \label{eq:scoring_function}
S(v_{\text{p}}, v_{\text{g}}) = 
\begin{cases} 
1, & \text{correct output} \\ 
0, & \text{incorrect or malformed output}
\end{cases}
\end{equation}




\paragraph{Format Reward.} 
We employ a binary format reward to enforce adherence to the prescribed output structure. This reward function assigns a value of $1.0$ if the model's response strictly follows the required template format, and $0$ otherwise. Specifically, the model must structure its output with two distinct components: (1) a reasoning process enclosed within \texttt{<thinking>...<\!/thinking>} tags, and (2) a final answer in JSON format enclosed within \texttt{<answer>```json...```<\!/answer>} tags. 

To implement this structured output format, we utilize carefully designed prompts during Chart-RL training as shown below. The system prompt establishes the conversational framework and defines the expected output structure, while the generation prompt reinforces these formatting requirements for each specific query.

\begin{center}
\begin{promptbox}[System Prompt]
A conversation between User and Assistant. The user asks a question, and the Assistant solves it. The assistant first thinks about the reasoning process in the mind and then provides the user with the answer. The reasoning process and answer are enclosed within <thinking> </thinking> and <answer> </answer> tags, respectively, i.e., <thinking> reasoning process here </thinking><answer> answer here </answer>
\end{promptbox}
\label{box:system_prompt}
\end{center}


\section{Experiments}

\begin{table*}[!t]
\normalsize
\centering
\begin{tabularx}{\textwidth}{@{}
    >{\centering\arraybackslash}m{2.5cm}  
    >{\centering\arraybackslash}m{1.5cm}  
    >{\centering\arraybackslash}X >{\centering\arraybackslash}X >{\centering\arraybackslash}X 
    >{\centering\arraybackslash}X >{\centering\arraybackslash}X >{\centering\arraybackslash}X 
    @{}}
\toprule
\multirow{2}{=}{\centering\textbf{Method}} 
& \multirow{2}{=}{\centering\textbf{Training}\\\textbf{Task}}
& \multicolumn{3}{c}{\shortstack{\textbf{MultiChartQA} \\ {\fontsize{7}{8.5}\selectfont\textit{Overall}}}}
& \multicolumn{3}{c}{\shortstack{\textbf{ChartInsights} \\ {\fontsize{7}{8.5}\selectfont\textit{Reasoning}}}} \\
\cmidrule(lr){3-5} \cmidrule(lr){6-8}
 & & Accuracy & Rel. $\Delta$ & {$p$-value}
 & Accuracy & Rel. $\Delta$ & {$p$-value} \\
\midrule
Baseline   & -- & 35.2 & -- & -- & 36.4 &  --  & -- \\ 
\midrule
SFT        & \multirow{3}{=}{\centering Hard}  & \cellcolor{gray!10} {32.9} & -6.6\% & 0.12 & \cellcolor{gray!10} {34.7} & -4.7\% & $<10^{-2}$ \\
CoT-SFT    & & \cellcolor{gray!10} \underline{39.1} & +11.1\% & $<10^{-2}$ & \cellcolor{gray!10} \underline{40.3} & +10.7\% & $<10^{-6}$ \\
{Chart-RL(ours)}   & & \cellcolor{gray!10}\bf{41.1} & {+16.7\%} &  {$<10^{-4}$} & \cellcolor{gray!10}\bf{40.6} & {+11.5\%} & {$<10^{-6}$} \\
\bottomrule
\end{tabularx}
\caption{Evaluation results on two benchmarks comparing our Chart-RL approach against baseline, SFT, and CoT-SFT. The rightmost column for each dataset shows the proportional test results for the relative change.}
\label{tab:table_chart_eval}
\end{table*}

\subsection{Training \& Evaluation Setup}
\paragraph{Baseline VLM.} We use Qwen2.5-VL-3B-Instruct as our baseline. Recent studies show that small language models (SLMs) are sufficiently powerful and, when adapted with efficient post-training, provide a cost-effective and therefore more promising foundation for practical agentic systems \citep{belcak2025small}. In addition, SLMs align better with our practical development needs.

\paragraph{Training Datasets.}
We conduct RL training on two types of chart‐based tasks:
\begin{itemize}
  \item \textbf{Easy Task:} 
    \begin{itemize}
      \item This task requires only single-step reasoning and involves straightforward chart question-answering, such as direct numeric extraction from a single chart element.
      \item We use 6,200 subsampled charts from PlotQA \citep{methani2020plotqa}, focusing on straightforward numerical queries.
    \end{itemize}
  
  \item \textbf{Hard Task:} 
    \begin{itemize}
      \item This task involves multi-step reasoning by performing a sequence of simple subtasks across one or multiple charts, ultimately producing responses through the aggregation of intermediate results.
      \item From the CharXiv validation set \citep{wang2024charxiv}, we subsample 448 charts with mathematically verifiable ground truth, tailored for multi-hop reasoning in Chart-RL training.          
    \end{itemize}
\end{itemize}

Unlike the easy tasks that focus on straightforward numerical extraction, the hard tasks require multi-hop reasoning across chart elements, making the 448-chart training subset sufficiently complex for RL training. Unless otherwise noted, all reported Chart-RL results use query-chart pairs from this 448-chart hard task training set. We provide examples of easy and hard tasks in Appendix~\ref{appendix:easy_vs_hard_task}. 

\paragraph{Training Hyper-parameters.}
We employ the VLM-R1 \citep{shen2025vlm} pipeline using our customized accuracy and formatting rewards. All RL experiments are conducted on 8 NVIDIA H100 GPUs. At each update, we generate 8 samples per input prompt. We set per-device batch size to 1 and accumulate gradients over 4 steps. We apply parameter-efficient fine-tuning (PEFT) via LoRA \citep{hu2022lora}, setting the adapter with rank $r=64$, scaling factor $alpha=128$, and dropout rate of 0.05. To enable joint vision-language adaptation, we keep all vision modules unfrozen throughout RL fine-tuning. 

\paragraph{Evaluation Benchmarks.}  
We evaluate Chart-RL on three state-of-the-art chart comprehension benchmarks, including MultiChartQA \citep{zhu2024multichartqa}, ChartInsights \citep{wu2024chartinsights}, and RobustCQA \citep{mukhopadhyay2024unraveling}. In addition to our analysis of aggregate scores across all benchmarks, we perform categorical analysis to facilitate task-specific comparisons.
\begin{itemize}
  \item \textbf{RobustCQA}: Measures the robustness of VLMs under systematically perturbed chart variations. 
  \item \textbf{MultiChartQA}: Assesses the model’s ability to perform multi-hop reasoning and to integrate information across multiple charts.
  \item \textbf{ChartInsights}: Evaluates fine-grained analytical performance across seven chart types and ten task categories.
\end{itemize}

\begin{table}[ht]
\normalsize
\centering
\label{tab:benchmark_statistics}
\begin{tabularx}{\columnwidth}{>{\centering\arraybackslash}m{1.4cm} >{\centering\arraybackslash}m{1.8cm} >{\centering\arraybackslash}m{1.4cm} >{\centering\arraybackslash}m{1.6cm}}
\toprule
\textbf{Category} & \textbf{Dataset} & \textbf{\# Charts} & \textbf{\# Queries} \\
\midrule
\multirow{2}{*}{Training} & {CharXiv} & 448 & 448 \\
                          & {PlotQA} & 6,200 & 6,200 \\
\midrule
\multirow{3}{*}{Evaluation} & {RobustCQA} & 400 & 400 \\
                            & {MultiChartQA} & 1,370 & 2,000 \\
                            & {ChartInsights} & 2,000 & 22,000 \\
\bottomrule
\end{tabularx}
\caption{Statistics of training and evaluation datasets.}
\end{table}

\paragraph{CoT Trajectory Generation.} 
We use GPT-4o as a teacher model to process query-chart pairs from the hard task and generate validated reasoning traces. The resulting CoT trajectories then serve as supervised fine-tuning data. To compare CoT-SFT against RL training, the complete prompt template is provided in the Appendix~\ref{appendix:cot_prompt_template}.  

\subsection{Fine-Tuning Strategy Comparison}
\label{chart-main-eval}

We evaluate three fine-tuning approaches, ranging from standard SFT, chain-of-thought SFT (CoT-SFT), to our Chart-RL, against the baseline VLM. Table~\ref{tab:table_chart_eval} summarizes the results across two chart understanding benchmarks. Chart-RL significantly outperforms both SFT and CoT-SFT approaches. All reported results are statistically significant at $\alpha=0.05$, except for SFT on MultiChartQA (p=0.12), with statistical significance confirmed across all methods on the larger ChartInsights dataset. 

Specifically, SFT leads to performance regression on both datasets, demonstrating that task-agnostic instruction tuning fails to capture the complexities of chart question answering. CoT-SFT shows substantial improvements over both baseline and standard SFT, achieving +11.1\% and +10.7\% relative gains on MultiChartQA and ChartInsights, respectively. Notably, although Chart-RL's training is not optimized for these specific benchmarks, it demonstrates consistent generalization capabilities, achieving relative improvements of 16.7\% on MultiChartQA and 11.5\% on ChartInsights.

We attribute Chart-RL's generalization and reasoning capabilities to the exploration inherent in RL training \citep{peng2025lmm}. These results demonstrate that our Chart-RL framework provides a robust approach for advancing chart understanding tasks.

\subsection{Robustness Analysis}
We evaluate Chart-RL's robustness across 25 perturbed chart categories using complex charts and questions from the RobustCQA dataset, which systematically assesses VLMs' consistency across different visual representations of identical underlying data. As shown in Table~\ref{tab:robustness_comparison}, Chart-RL demonstrates improved performance in 18 of 25 categories (72\%), compared to the SFT. In contrast, SFT only achieves better results in only 2 categories (8\%). The remaining 5 categories exhibits identical performance (20\%).

The most significant improvements occur in categories involving chart layout modifications and visual styling, such as hatching patterns, legend positioning, and tick orientation. These improvements suggest that reinforcement learning with chart-specific reward signals likely enhances generalization capabilities, making the model robust to layout and structural variations. Interestingly, Chart-RL shows minor degradation in only two scenarios: log scale transformations and scaling perturbations. This observation suggests that mathematical transformations may require additional reasoning patterns during training.    

\begin{table}[ht!]
\normalsize
\centering
\setlength{\tabcolsep}{6pt}
\begin{tabular}{l | cc}
\toprule
\multirow{2}{*}{\textbf{Category}} &
  \multicolumn{2}{c}{\textbf{Fine-Tuning Strategy}} \\
\cmidrule(lr){2-3}
  & SFT & Chart‑RL (ours) \\
\midrule
annotations & \underline{0.41} & \underline{0.41} \\
benchmark & 0.47 & \textbf{0.52} \\
\makecell[l]{color\_random} & \underline{0.37} & \underline{0.37} \\
\makecell[l]{color\_scheme} & \underline{0.34} & \underline{0.34} \\
\makecell[l]{data\_pivot} & 0.35 & \textbf{0.36} \\
font & 0.39 & \textbf{0.41} \\
grid & 0.38 & \textbf{0.41} \\
hatching & 0.28 & \textbf{0.34} \\
\makecell[l]{horizontal\_grouped} & 0.35 & \textbf{0.38} \\
\makecell[l]{horizontal\_stacked} & 0.34 & \textbf{0.37} \\
\makecell[l]{legend\_position} & 0.34 & \textbf{0.40} \\
\makecell[l]{line\_representation} & 0.32 & \textbf{0.35} \\
log\_scale & \textbf{0.33} & 0.30 \\
normal & 0.37 & \textbf{0.41} \\
\makecell[l]{only\_data\_color\_scheme} & \underline{0.35} & \underline{0.35} \\
\makecell[l]{replacing\_legend\_with\_labels} & 0.35 & \textbf{0.36} \\
scaling & \textbf{0.34} & 0.32 \\
\makecell[l]{scatter\_representation} & \underline{0.31} & \underline{0.31} \\
stacked & 0.30 & \textbf{0.32} \\
stacked\_area & 0.32 & \textbf{0.33} \\
\makecell[l]{stair\_plot\_normal} & 0.23 & \textbf{0.29} \\
\makecell[l]{stair\_plot\_with\_marker} & 0.32 & \textbf{0.35} \\
stem\_plot & 0.30 & \textbf{0.31} \\
\makecell[l]{tick\_orientation} & 0.37 & \textbf{0.43} \\
tick\_position & 0.36 & \textbf{0.37} \\
\bottomrule
\end{tabular}
\caption{Robustness comparison between Chart-RL and SFT across 25 perturbed chart categories.}
\label{tab:robustness_comparison}
\end{table}

\subsection{Data Efficiency in RL Training} 

\begin{table*}[!ht]
\normalsize
\centering
\begin{tabularx}{\textwidth}{@{}
    >{\centering\arraybackslash}m{2.2cm} 
    >{\centering\arraybackslash}m{1.2cm} 
    >{\centering\arraybackslash}m{1.2cm} 
    >{\centering\arraybackslash}X >{\centering\arraybackslash}X >{\centering\arraybackslash}X 
    >{\centering\arraybackslash}X >{\centering\arraybackslash}X >{\centering\arraybackslash}X 
    @{}}
\toprule
\multirow{2}{=}{\centering\textbf{Method}} 
& \multirow{2}{=}{\centering\textbf{Training}\\\textbf{Task}}
& \multirow{2}{*}{\makecell[c]{\textbf{Training}\\\textbf{Samples}}} 
& \multicolumn{3}{c}{\shortstack{\textbf{MultiChartQA} \\ {\fontsize{7}{8.5}\selectfont\textit{Overall}}}}
& \multicolumn{3}{c}{\shortstack{\textbf{ChartInsights} \\ {\fontsize{7}{8.5}\selectfont\textit{Reasoning}}}} \\
\cmidrule(lr){4-6} \cmidrule(lr){7-9}
 & & & Accuracy & Rel. $\Delta$ & {$p$-value}
 & Accuracy & Rel. $\Delta$ & {$p$-value} \\
\midrule
Baseline   & -- & --   & {35.2}   & --      & -- & {36.4}   & --      & -- \\
\midrule
\multirow{3}{*}{Chart-RL (ours)} 
& \multirow{3}{=}{\centering Hard}
& 10   & \cellcolor{gray!10} {39.1}   & +11.1\% & $<10^{-2}$ & \cellcolor{gray!10} {39.1}   & +7.4\%  & $<10^{-4}$ \\
& & 100  & \cellcolor{gray!10} \underline{39.2}   & +11.4\% & $<10^{-2}$ & \cellcolor{gray!10} \underline{39.7}   & +9.1\%  & $<10^{-6}$ \\
& & 448  & \cellcolor{gray!10} \bf{41.1}   & +16.7\% & $<10^{-4}$ & \cellcolor{gray!10} \bf{40.6}   & +11.5\% & $<10^{-6}$ \\
\bottomrule
\end{tabularx}
\caption{Evaluation results for Chart-RL across different training sample sizes, showing performance on three chart understanding benchmarks under varying RL training data scales. The rightmost column for each dataset shows the proportional test results for the relative change.}
\label{tab:chart_eval_training_efficiency_1}
\end{table*}

\begin{figure*}[!t]
    \centering
    \includegraphics[width=0.9\linewidth]{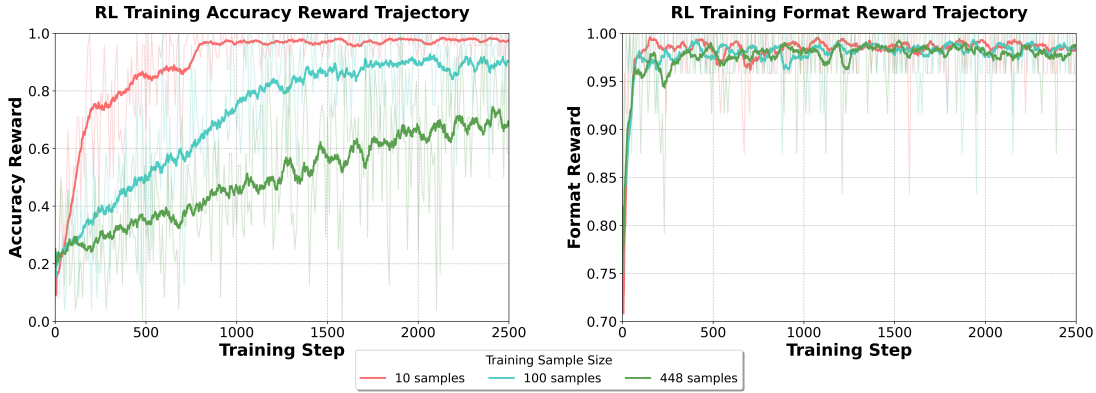}
    \caption{RL Training reward trajectories for Chart-RL across different training sample sizes, showing accuracy rewards (left) and format rewards (right) throughout the training process.}
    \label{fig:data_efficiency}
\end{figure*}

Inspired by recent work on RL training efficiency \citep{wang2025reinforcement}, we examine the training dynamics using minimal training examples. Specifically, we aim to investigate how few verifiable-reward samples are required to achieve efficient visual reasoning in chart. Figure~\ref{fig:data_efficiency} compares Chart-RL's training dynamics with limited training data, where the number of charts is restricted to 10, 100, or 448 randomly subsampled instances of equivalent task complexity. Notably, with only 10 samples, the accuracy reward converges rapidly and plateaus earlier than the 488-sample run. In addition, the format reward follows a similar convergence pattern across all sample sizes, likely reflecting the robust structural adherence capability of the Qwen2.5-VL instruct model. 

We extensively evaluate the resulting model checkpoints across three chart-related benchmarks. As shown in Table~\ref{tab:chart_eval_training_efficiency_1}, Chart-RL produces substantial improvements on MultiChartQA and ChartInsights despite using far fewer training samples. 

In summary, Chart-RL demonstrates strong generalization across diverse benchmarks while requiring only minimal training examples. Unlike traditional SFT, this data-efficient training paradigm enables effective adaptation to both novel chart tasks and predefined output formats.

\subsection{Out-of-Domain Generalization} 
We investigate whether Chart-RL's reasoning capabilities generalize to out-of-domain visual mathematical tasks. As shown in Table~\ref{tab:table_ood_eval}, Chart-RL achieves a significant 55.6\% relative improvement on MathVerse \citep{zhang2024mathverse}, despite not being explicitly trained on mathematical reasoning data. These results demonstrate that Chart-RL effectively transfers to visually grounded mathematical problem solving. 

Our findings align with recent studies highlighting domain-specific reinforcement learning can incentivize transferable analytical skills~\citep{xie2025logic}. We believe that complex chart-based reasoning represents an important yet underexplored training data source that could benefit the development of generalized VLMs \citep{liang2024survey}.

\begin{table}[!ht]
\normalsize
\centering
\begin{tabularx}{\columnwidth}{@{}
    >{\centering\arraybackslash}m{2.6cm} | 
    >{\centering\arraybackslash}X >{\centering\arraybackslash}X >{\centering\arraybackslash}X 
    @{}}
\toprule
\multirow{2}{*}{\makecell{\textbf{Methods}\\}} 
& \multicolumn{3}{c}{
    \makecell{
        \textbf{MathVerse} \\
        {\fontsize{7}{8.5}\selectfont\textit{Vision}}
    }
} \\
\cmidrule(lr){2-4}
 & \makecell{Acc.} & \makecell{Rel. $\Delta$} & \makecell{$p$-value} \\
\midrule
Baseline   & \cellcolor{gray!10} {1.8} & -- & -- \\ 
Chart-RL (ours)   & \cellcolor{gray!10}\bf{2.8} & +55.6\% & $<10^{-2}$ \\
\bottomrule
\end{tabularx}

\caption{Evaluation results on MathVerse Vision.}
\label{tab:table_ood_eval}
\end{table}
\section{Ablation Study}
\subsection{Generalization in Low-Level Tasks} 
\begin{figure}[!t]
    \centering
    \includegraphics[width=0.85\linewidth]{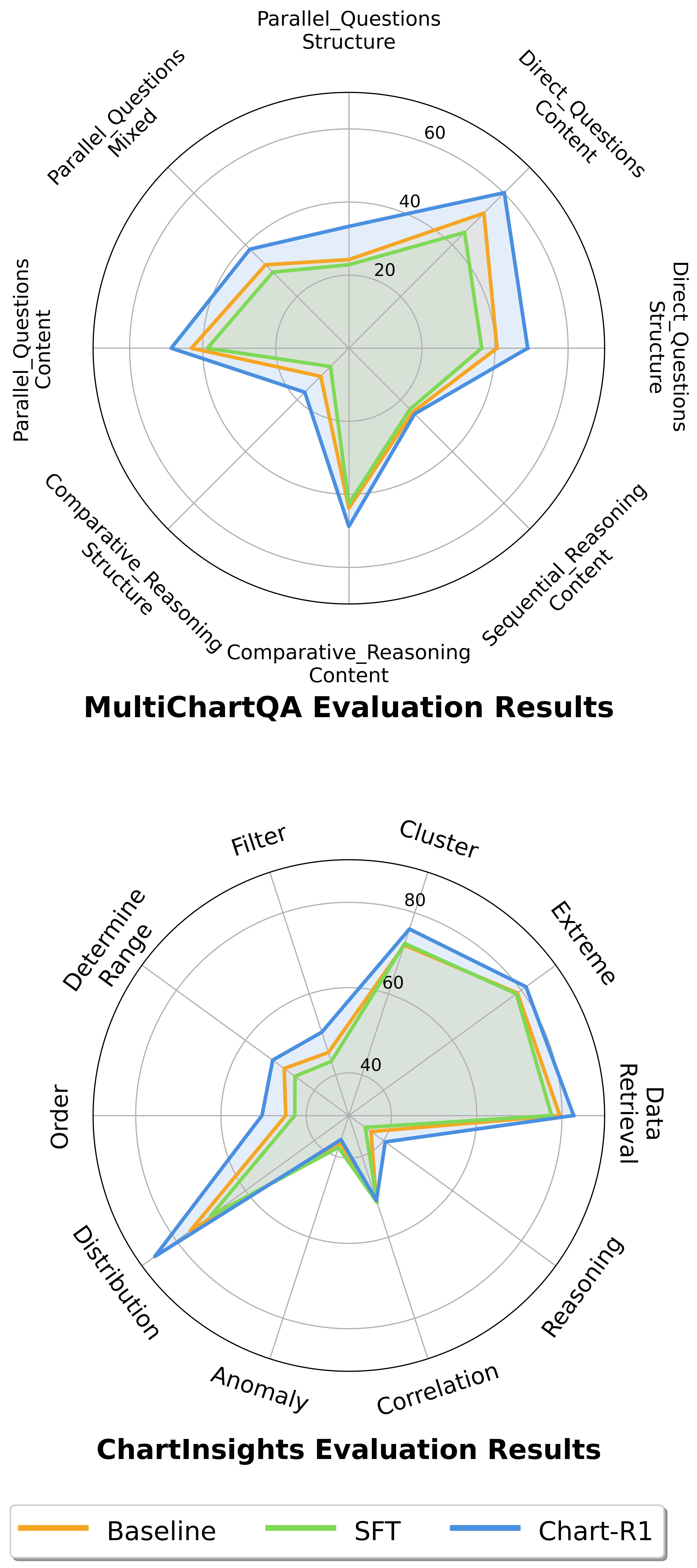}
    \caption{Performance breakdown by sub-category on MultiChartQA and ChartInsights benchmarks.}
    \label{fig:abalation_eval_res}
\end{figure}

To evaluate Chart-RL's generalization capabilities within the chart understanding domain, we compare Chart-RL against the SFT-only variant and the base Qwen2.5-VL-3B-Instruct model. Figures \ref{fig:abalation_eval_res} presents detailed per-category performance comparisons on ChartInsights and MultiChartQA benchmarks.

Chart-RL consistently outperforms both the baseline and the SFT-enhanced variant across various low-level tasks. Our method demonstrates robust improvements, likely resulting from the reinforced exploration on challenging chart-query pairs during RL training. Specifically, Chart-RL achieves substantial gains in reasoning tasks (Direct Questions, Parallel Questions, and Comparative Reasoning) on MultiChartQA, and data operations  (Order, Filter, and Cluster) on ChartInsights. 

By training on highly complex chart samples, we enable effective transfer to simpler, low-level chart-related tasks without requiring additional task-specific training. This suggests that our method facilitates fundamental chart understanding capabilities that generalize beyond the predefined training tasks.

\subsection{Task Complexity Matters: Easy vs. Hard Training Tasks} 
\label{sec:training-task-complexity}

\begin{figure}[!ht]
    \centering
    \includegraphics[width=1.0\linewidth]{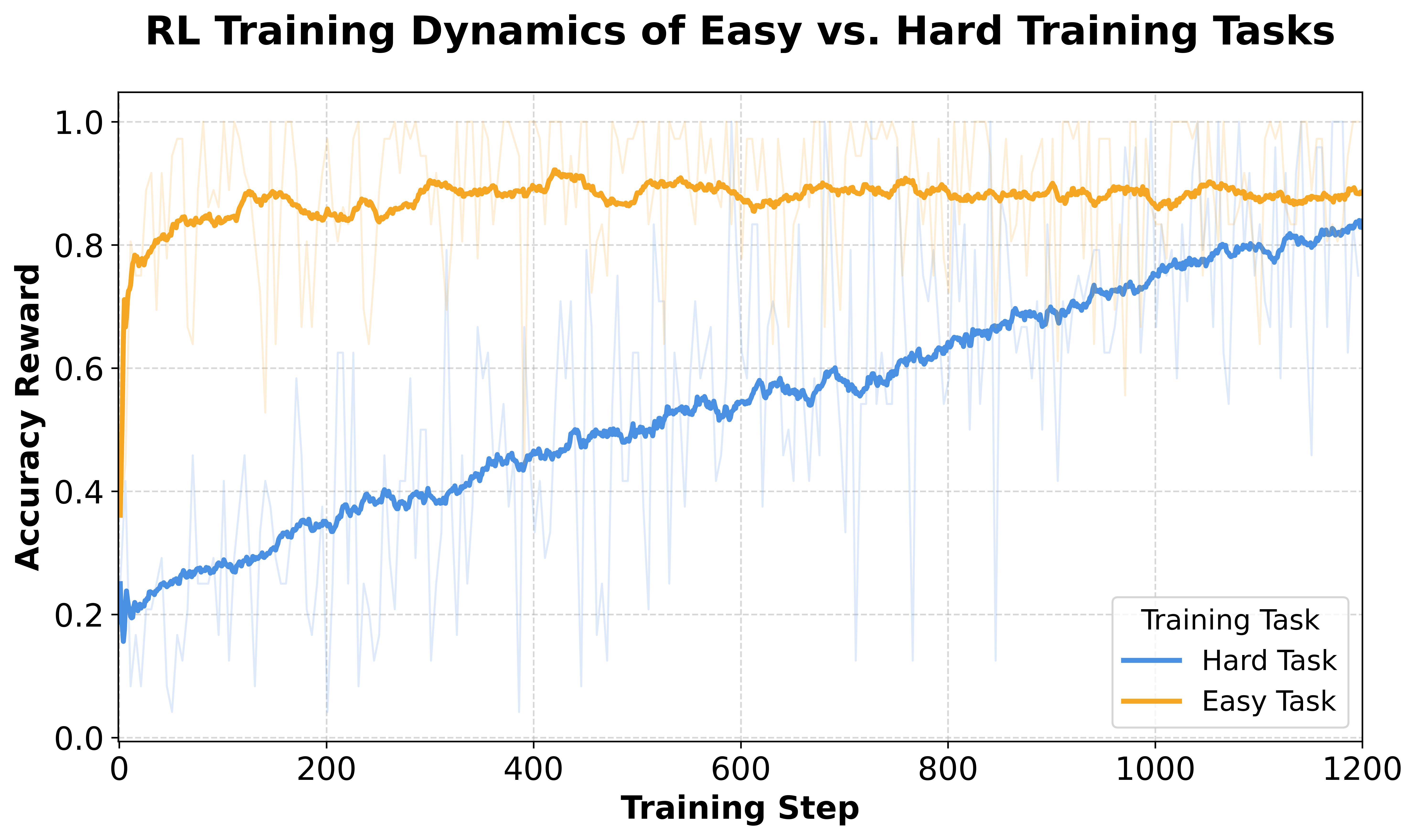}
    \caption{Accuracy reward trajectories in RL training for Easy and Hard tasks.}
    \label{fig:easy_vs_hard}
\end{figure}

To investigate how the complexity of training tasks affects RL performance, we compare models trained on tasks of varying difficulty. Figure~\ref{fig:easy_vs_hard} shows drastically different learning dynamics. Specifically, easy tasks achieve high accuracy rapidly but offer limited learning potential, plateauing around {0.9} with minimal subsequent improvement. In contrast, hard tasks begin with lower accuracy of {0.2} but demonstrate sustained improvement throughout RL training. This continuous learning trajectory suggests that hard tasks provide helpful feedback signals, enabling the model to develop more sophisticated chart reasoning capabilities.

Table~\ref{tab:task-difficulty} summarizes the impact of task complexity on downstream performance. While easy tasks attain higher training accuracy, it ultimately leads to degradation on both evaluation benchmarks. By contrast, hard tasks, though achieving relatively lower training accuracy, result in substantial improvements across all benchmarks. These results highlight a critical insight: training on challenging tasks that maintain a consistent learning signal throughout optimization leads to better generalization. The continuous improvement on hard tasks suggests that the model develops robust chart comprehension capabilities, while the early saturation on easy tasks likely leads to overfitting on limited patterns. Our results indicate that the complexity and diversity of training tasks are essential in RL for developing generalizable chart comprehension capabilities. 



\begin{table}[H]
\normalsize  
\centering
\newcommand{\reldelta}[1]{{\scalebox{0.8}{\,(#1)\,}}}
\begin{tabularx}{\columnwidth}{l | *{2}{>{\centering\arraybackslash}X}}
\toprule
\textbf{Task} & \textbf{MultiChartQA} & \textbf{ChartInsights} \\
\midrule
Baseline & 35.2 & 36.4 \\
\midrule
Easy & 34.8\,\reldelta{-1.2\%} & 36.0\,\reldelta{-1.1\%} \\
Hard & \cellcolor{gray!10} \textbf{41.1}\,\reldelta{+16.8\%} & \cellcolor{gray!10} \textbf{40.6}\,\reldelta{+11.5\%} \\
\bottomrule
\end{tabularx}
\caption{Evaluation results of Easy vs. Hard training tasks on different benchmarks.}
\label{tab:task-difficulty}
\end{table}
\section{Conclusion}
We present Chart-RL, an effective RL-based method that significantly advances task-agnostic chart comprehension capabilities in VLMs. Through systematic experiments across diverse benchmarks, we demonstrate that Chart-RL achieves substantial improvements using only a few hundreds of training samples. Our analysis reveals that training on complex tasks that provide continuous learning signals produces better generalization compared to optimizing for high training accuracy on simpler tasks. Moreover, Chart-RL exhibits cross-domain transfer capability, improving multimodal mathematical reasoning performance without explicit training in this domain.


\section*{Limitations}
Our work has a few limitations that should be acknowledged. First, Chart-RL relies on mathematically verifiable ground truths for reward computation, which restricts its applicability to chart understanding tasks that lack deterministic answers or require subjective interpretation.  Second, the evaluation remains relatively narrow and may not capture the full diversity of real-world chart comprehension scenarios. The Chart-RL training requires careful curation of complex chart reasoning tasks, which may not be readily available for all domains or chart types. Third, our work mainly focus on the initial application of RLVR for chart question answering. We believe that a multi-stage post-training strategy, alternating SFT and RL, might lead to enhanced accuracy and generalization. 



\bibliography{custom}

\clearpage
\appendix
\section{Appendix}
\label{sec:appendix}





\subsection{CoT Prompt Template} 
\label{appendix:cot_prompt_template}
We use the GPT-4o model via Azure OpenAI (API version 2024-12-01-preview) to generate explicit chain-of-thought (CoT) reasoning. Below is the prompt template:

\begin{center}
\begin{promptbox}[Prompt for Generating CoT Trajectories with GPT-4o]
I will give you an image, an original question, and its correct answer. Your task is to rewrite the question so that solving it requires step-by-step CoT reasoning, including any necessary numerical or mathematical expressions. Feel free to include natural thought markers (e.g., ‘let me think,’ ‘oh, I see’).

\medskip
• The rewritten question must request a specific, easily verifiable answer (for example, ‘2’ or ‘A’)—no open-ended prompts.  
• Do not include the word ‘Answer:’ in the question to avoid leakage.

\medskip
\textbf{Input:}  
Original Question: \{original\_question\}  
Original Answer: \{original\_answer\}

\medskip
\textbf{Output:}  
Question: [Rewritten question]  
Answer: [Detailed reasoning steps, with calculations as needed]  
<think>Step-by-step reasoning process</think>  
<answer>Final, verifiable answer</answer>
\end{promptbox}
\end{center}

\subsection{Examples of Easy and Hard Tasks}
\label{appendix:easy_vs_hard_task}
Easy tasks typically involve direct information extraction where the answer can be found through simple visual inspection, such as reading a single value from a clearly labeled data point. In contrast, hard tasks require multi-step reasoning, including comparing multiple data points, identifying relationships between variables, and performing intermediate calculations or rankings.

\begin{figure}[ht]
    \centering
    \includegraphics[width=0.8\linewidth]{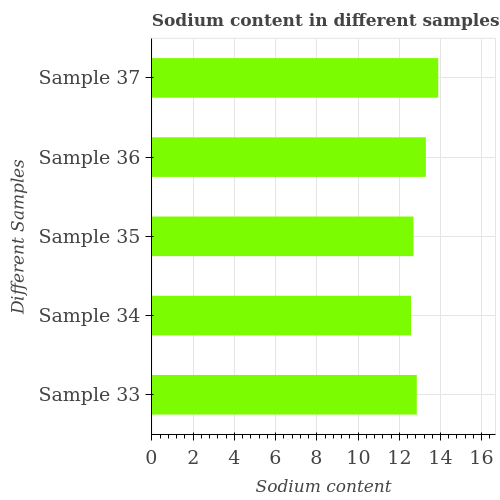}
    
    \caption{An example of easy task for RL training.}
    \label{fig:easy_task}
\end{figure}

The query for the easy task in Figure~\ref{fig:easy_task} is: ``What is the sodium content of sample 34?'' The corresponding ground truth annotation is 12.8.

The query for the hard task in Figure~\ref{fig:hard_task} is: ``Provide the accuracy (\%) of the model with the second highest time.'' The corresponding ground truth annotation is 94.

\begin{figure}[ht]
    \centering
    \includegraphics[width=0.8\linewidth]{figs/simple_task.jpg}
    \caption{An example of hard task for RL training.}
    \label{fig:hard_task}
\end{figure}

The corresponding chain-of-thought (CoT) reasoning trajectory from GPT-4o is shown below:

\begin{center}
\begin{promptbox}[CoT Reasoning Trajectory for the Hard Example]
\texttt{<think>}To determine the accuracy of the model with the second highest execution time, we \textcolor{red}{first} observe the x-axis labeled ``Time (s)'' and identify the models based on their execution times. The longest time corresponds to the model Xception at approximately 6 seconds. The second longest time corresponds to DenseNet201 at approximately 5 seconds.  \textcolor{red}{Next}, we look at the y-axis labeled ``Accuracy (\%)'' to find the accuracy value for DenseNet201. The accuracy for DenseNet201 is clearly marked as 94\%.\texttt{</think>}

\texttt{<answer>}94\texttt{</answer>}
\end{promptbox}
\end{center}

\end{document}